\documentclass[letterpaper, 10 pt, conference]{ieeeconf}  

\IEEEoverridecommandlockouts                              

\let\labelindent\relax   
\usepackage{enumitem}
\usepackage{graphics} 
\usepackage{graphicx}
\usepackage{amsmath} 
\usepackage{amssymb}  
\usepackage{amsfonts}
\usepackage{booktabs}
\usepackage[table]{xcolor}   
\usepackage{multirow}
\usepackage{siunitx}
\usepackage{makecell}
\usepackage{cite}
\usepackage{subcaption}

\usepackage[dvipsnames]{xcolor}
\usepackage[colorlinks=true,
            linkcolor=cvprblue,
            citecolor=cvprblue,
            filecolor=cvprblue]{hyperref}

\definecolor{cvprblue}{rgb}{0.21,0.49,0.74}
\definecolor{myblue}{rgb}{0.88,0.98,1} 
\definecolor{deemph}{gray}{0.6}

\makeatletter
\def\blfootnote{\xdef\@thefnmark{}\@footnotetext}
\makeatother

\title{\LARGE \bf
From Power to Precision: Learning Fine-grained Dexterity \\ for Multi-fingered Robotic Hands
}

\author{\authorblockN{Jianglong Ye$^*$, Lai Wei$^*$, Guangqi Jiang, Changwei Jing\\Xueyan Zou, Xiaolong Wang}
\vspace{0.05in}
\authorblockA{UC San Diego}
{\href{https://jianglongye.com/power-to-precision}{\texttt{jianglongye.com/power-to-precision}}}
}

\begin{document}

\twocolumn[{%
\renewcommand\twocolumn[1][]{#1}%
\maketitle
\begin{center}
    \vspace{-1.5em}
    \centering
    \captionsetup{type=figure}
    \includegraphics[width=\linewidth]{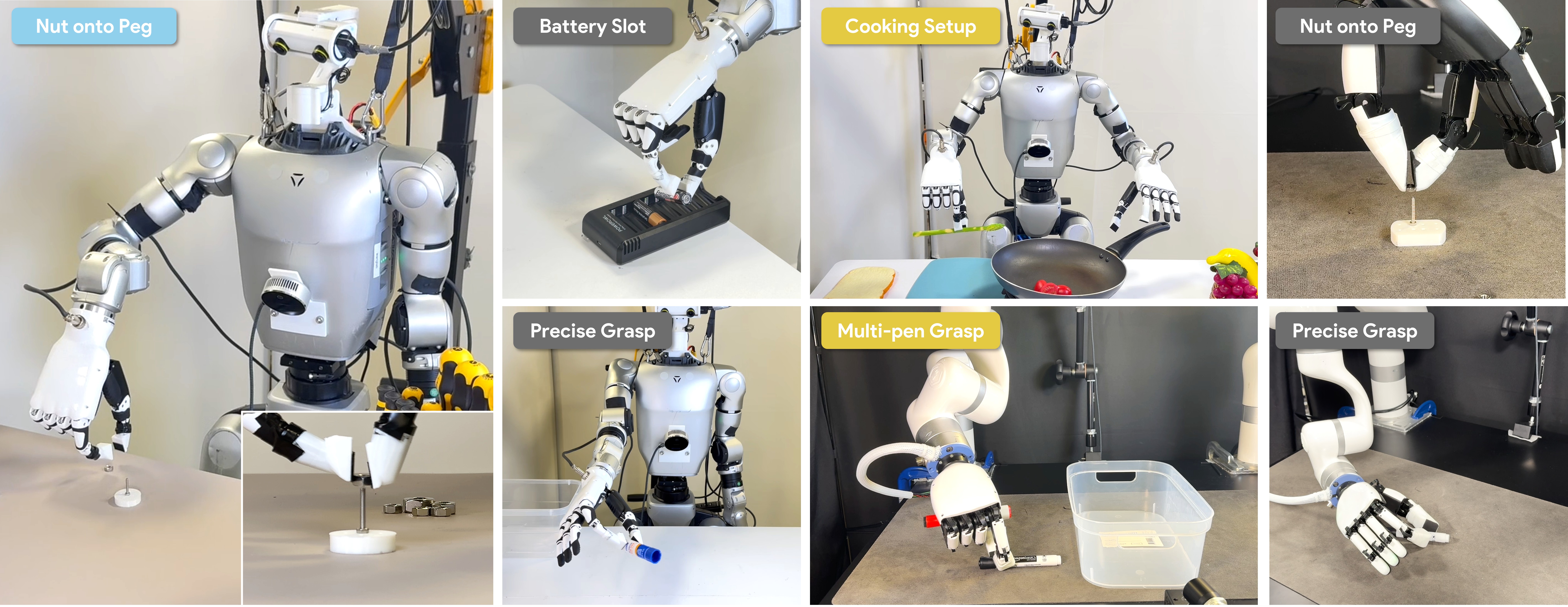}
    \vspace{-1.5em}
    \caption{Our system enables multi-fingered robotic hands to perform a diverse set of manipulation tasks, ranging from precise manipulation of small objects (e.g., pens, nuts, batteries) to power grasps of larger items (e.g., frying pans). This versatility and dexterity are achieved through a co-design framework for both control and fingertip-geometry optimization.}
    \label{fig:teaser}
\end{center}
}]
{\blfootnote{{$^{*}$ Equal contribution.}}}

\thispagestyle{empty}
\pagestyle{empty}

\begin{abstract}
Human grasps can be roughly categorized into two types: power grasps and precision grasps. Precision grasping enables tool use and is believed to have influenced human evolution. Today's multi-fingered robotic hands are effective in power grasps, but for tasks requiring precision, parallel grippers are still more widely adopted.
This contrast highlights a key limitation in current robotic hand design: the difficulty of achieving both stable power grasps and precise, fine-grained manipulation within a single, versatile system.
In this work, we bridge this gap by jointly optimizing the control and hardware design of a multi-fingered dexterous hand, enabling both power and precision manipulation. Rather than redesigning the entire hand, we introduce a lightweight fingertip geometry modification, represent it as a contact plane, and jointly optimize its parameters along with the corresponding control.
Our control strategy dynamically switches between power and precision manipulation and simplifies precision control into parallel thumb–index motions, which proves robust for sim-to-real transfer. On the design side, we leverage large-scale simulation to optimize the fingertip geometry using a differentiable neural-physics surrogate model.
We validate our approach through extensive experiments in both sim-to-real and real-to-real settings. Our method achieves an 82.5\% zero-shot success rate on unseen objects in sim-to-real precision grasping, and a 93.3\% success rate in challenging real-world tasks involving bread pinching. These results demonstrate that our co-design framework can significantly enhance the fine-grained manipulation ability of multi-fingered hands without reducing their ability for power grasps.
\end{abstract}

\section{Introduction}
A classical taxonomy divides human grasps into two main categories: the \textit{power grasp}, which secures objects against the palm with the fingers, and the \textit{precision grasp}, which places the thumb in opposition to other fingertips~\cite{napier1956prehensile,feix2015grasp}.
While both are fundamental, the precision grasp is particularly associated with the fine-grained manipulation required for tool use in early humans, which significantly shaped the evolutionary trajectory of our species~\cite{skinner2015human,kivell2015evidence,karakostis2021biomechanics}.

Recent advancements have shown that multi-fingered robotic hands are effective in power grasp, as more contact points provide greater stability~\cite{ye2025dex1b,chen2025bodex,singh2024dextrah,fang2025anydexgrasp}. In contrast, for fine-grained manipulation tasks requiring precision, two-finger parallel grippers are more widely adopted, with impressive applications ranging from folding shirts to gear insertion~\cite{zhao2023learning,zhao2024aloha,black2024pi_0,intelligence2025pi05,team2025gemini}. Therefore, replicating the human-level dexterity in multi-fingered hands, especially for precision-oriented manipulation, remains a fundamental challenge.

Our goal is to enable a multi-fingered dexterous hand with reliable precision manipulation while preserving strong power grasp capability. We aim for the following goals: (i) plug-and-play compatibility with existing commercially available multi-fingered hands by augmenting fingertip geometry rather than designing a new hand from scratch, (ii) achieving both power and precision manipulation in a unified hardware platform and control strategy, and (iii) validation with both sim-to-real transfer using large-scale simulation and real-to-real policies using teleoperation.

To achieve these goals, we adopt a joint control–design optimization framework. On the control side, we switch between power and precise manipulation, where precise manipulation is simplified into coordinated thumb–index actions and parallel finger motions. We also propose a neural switcher that dynamically switch to corresponding grasp mode based on the object. On the design side, we leverage large-scale simulation with a neural physics surrogate model to co-optimize fingertip geometry and control variables, yielding physically effective fingertip designs.
Together, these components provide a simple and generalizable solution for both precise and power manipulation.

We validate the proposed system in both sim-to-real and real-to-real settings. In the sim-to-real setting, we focus on grasping tasks, where objects are categorized into power grasps and precise grasps. Demonstrations are optimized with multiple objectives, and a policy is learned from them. Real-world experiment results show that our method outperforms the SOTA method on precise grasps by a large margin. In the real-to-real setting, we tackle more difficult compositional task and precision task. Demonstrations are collected using our switchable teleoperation, where precise manipulation use optimized control. The trained policies are capable of picking up an M4 hex nut with precision and power grasping a pan handle, as shown in Fig.~\ref{fig:teaser}.

In summary, we claim the following contributions:
\begin{itemize}[leftmargin=*]
\item Control optimization for precise manipulation, including thumb-index motion generation for both grasp synthesis and real-world teleoperation.
\item Design optimization with a contact-plane geometry representation and a neural-physics surrogate distilled from large-scale simulations.
\item Extensive experiments that demonstrate the effectiveness of our co-design framework in both simulation and the real world, showing that our system improves precise manipulation ability without compromising power grasp.
\end{itemize}

\section{Related Work}

\textbf{Power to Precise Manipulation.} 
\textit{Power manipulation} refers to using whole-hand contacts to apply large, robust forces that move or stabilize objects without requiring high pose accuracy (e.g., holding a hammer, grasping a large bottle). Recent studies~\cite{ye2025dex1b,chen2025bodex,zhang2024graspxl,zhong2025dexgraspvla,he2025dexvlg,singh2024dextrah,fang2025anydexgrasp,cheng2024open,ding2024bunny} have shown that multi-fingered robotic hands perform well in power manipulation, mainly because additional contact points increase stability.
For example,~\cite{ye2025dex1b,chen2025bodex} leverage large-scale simulation to learn dexterous grasping policies and employ point cloud observations for robust sim-to-real transfer. ~\cite{cheng2024open,ding2024bunny} use VR devices to collect high-quality teleoperation data and demonstrate impressive dexterous hand tasks.

In contrast, \textit{precise manipulation} refers to using fingertip-scale, finely controlled motions and contacts to achieve accurate object poses and delicate interactions (e.g., inserting a key, turning a screw). For these precise tasks, parallel grippers are more widely adopted~\cite{zhao2023learning,zhao2024aloha,black2024pi_0,intelligence2025pi05,team2025gemini}. How to achieve precise manipulation, especially with multi-fingered hands, is still an open problem for the community. To this end, previous works~\cite{zhao2023gelsight,hong2023angle,do2024densetact,yu2023precise,bronars2024texterity} either design better hardware or use more accurate sensors such as tactile sensors. ~\cite{zhao2023gelsight,hong2023angle} optimize hardware design to achieve precise manipulation, while the more general approaches~\cite{do2024densetact,yu2023precise,bronars2024texterity} employ tactile sensing to facilitate precise manipulation. Instead of designing new hardware or using tactile sensing, our approach optimizes both control and finger tip geometry for precise manipulation.

\textbf{Computational Design and Co-Design.}
Previous computational methods for mechanical design often use gradient-based optimization and task-aware planning over geometry or topology to create structures for target tasks~\cite{chen2020design,feshbach2024algorithmic}. However, these pipelines can be computationally expensive and may not generalize well to new objects or tasks. Differentiable simulators address this by enabling the joint optimization of morphology and control within a single framework~\cite{calandra2016bayesian}. Recent work has increasingly utilized co-design as a tool for manipulation, such as in the creation of dexterous hands~\cite{mannam2024design}, fingertips~\cite{ikemuraefficient}, and grippers~\cite{liu2024paperbot, yi2025co}. 
Rather than designing a completely new dexterous hand, we enhance an existing hand by optimizing its geometry. In our co-design framework, this is achieved by attaching a newly optimized fingertip cover that improves precise manipulation. During optimization, we model the fingertip geometry as a contact plane and leverage large-scale simulation by using a neural dynamics surrogate.
\textbf{Imitation Learning.} Imitation learning is a common framework for learning robot control policies from demonstrations, and it has been applied in many recent works~\cite{finn2017one,mandlekar2020learning,zhu2022bottom,nasiriany2022learning,mandlekar2021matters,qin2022dexmv,mandlekar2023mimicgen,ye2023learning,young2021visual,cheng2024open,ding2024bunny, qiu-song-peng-2024-wildlma, wei2024ensuringforcesafetyvisionguided}. Demonstrations may be obtained through teleoperation~\cite{mandlekar2018roboturk,brohan2022rt,wu2024gello,ding2024bunny}, simulation~\cite{wang2023robogen,james2020rlbench,wang2023dexgraspnet,ye2025dex1b}, real-world policy trial-and-error~\cite{pinto2016supersizing,kalashnikov2018scalable}, human videos~\cite{qin2022dexmv,ye2023learning,bahl2023affordances}, or a mixture of these sources~\cite{mandlekar2023mimicgen,dass2022pato,o2024open}. In this work, we validate our approach on real-to-real tasks using teleoperation data and on sim-to-real tasks using simulation data. We apply existing imitation learning methods~\cite{zhao2023learning,ye2025dex1b} to our collected demonstrations. Our co-design framework improves the robustness of both demonstrations and the robotic hand, leading to more effective precise manipulation policies in simulation and the real world.

\section{Method}
\label{sec:method}

\subsection{Overview}
This work focuses on learning manipulation skills for multi-fingered hands using imitation learning. In this framework, a control policy $\pi$ learns from expert demonstrations $D = (\mathcal{O}, \mathcal{A})$, consisting of observations $\mathcal{O}$ and actions $\mathcal{A}$ collected through teleoperation, simulation, or other methods. 
To achieve precise manipulation, we introduce an approach that jointly optimizes both control (Sec.~\ref{subsec:control}) and design (Sec.~\ref{subsec:design}) of dexterous hands. 
We cross validate our approach on tasks under both sim-to-real and real-to-real settings, where data is collected through optimization in simulation and teleoperation in the real world.

\textbf{Control Optimization.} Our control method follows two principles. The first is \textit{categorization}, where we switch between power and precision grasps based on the target object. For example, in our \textit{Cooking Setup} task, a power grasp is used for a pot handle, while a precision pinch grasp is used for a thin asparagus. The second principle is \textit{simplification}. We achieve this by reducing the degrees of freedom of the hand and the number of contact points. We also constrain finger motion to be parallel movements during precision grasps. Although it is possible to use more complex motions for precise manipulation, our experiments show that simpler motions are effective for both retargeting and sim-to-real transfer.

\textbf{Design Optimization.} An effective control strategy can still be limited by the hand's physical design.
To address this, we apply a co-design framework to optimize the fingertip geometry for precise manipulation. We model the contact region between the thumb and index finger as a plane and search for the optimal plane. 
This optimization process uses a learned, differentiable forward dynamics model along with several objectives, which allows us to leverage large-scale simulation to find a better fingertip design.

\subsection{Control Optimization for Precise Manipulation}
\label{subsec:control}


The robot’s action space $\mathcal{A}$ is defined as the target joint positions $q \in \mathbb{R}^d$, where $d = 19$ (7 for the arm and 12 for the hand). The collected data, in both simulation and the real world, consist of observations $\mathcal{O}$ and trajectories of actions $\{q_t\}_{t=1}^T$, where each $q_t \in \mathcal{A}$.

\textbf{Sim-to-real Grasping.} We proof the effectiveness of the proposed optimization pipeline in Sim-to-real settings. In simulation, a common way to collect demonstrations for dexterous grasping is to first optimize for force closure and then apply motion planning combined with simulation filtering~\cite{ye2025dex1b,wang2023dexgraspnet,chen2025bodex}. 
Although this framework can produce grasps that work even for tiny objects in simulation, the poses are often complex because all fingers are encouraged to apply force to the object. Deploying these grasps to the real world is impractical due to the sim-to-real gap.

We categorize grasping poses based on the type of target object. For simplicity, we focus on two categories: power grasp for large objects, and precision grasp for small or thin objects. For power grasp optimization, we use the original force closure objective from ~\cite{liu2021synthesizing}. For precision grasp, we introduce a variant objective. 
Given object mesh $O$ and joint configuration $q$, we sample $n$ contacts on the thumb and $n$ on the index fingertip. Each contact has position $x_i\in\mathbb{R}^3$ and normal $c_i\in\mathbb{R}^3$. The grasp map is
$
G=\begin{bmatrix}
I_3 & \cdots & I_3\\
[x_1]_{\times} & \cdots & [x_{2n}]_{\times}
\end{bmatrix},
$
where $[x]_{\times}$ is the skew-symmetric matrix of $x$. We minimize
$
E_{\mathrm{precise}}=\|G\,c\|_2,
$
with $c=[c_1^\top,\ldots,c_{2n}^\top]^\top$, which encourages the net wrench to approach zero for the thumb-index grasp. We also fix other fingers during optimization to reduce the degrees of freedom.

\begin{figure}[t]
    \centering
    \includegraphics[width=0.48\textwidth]{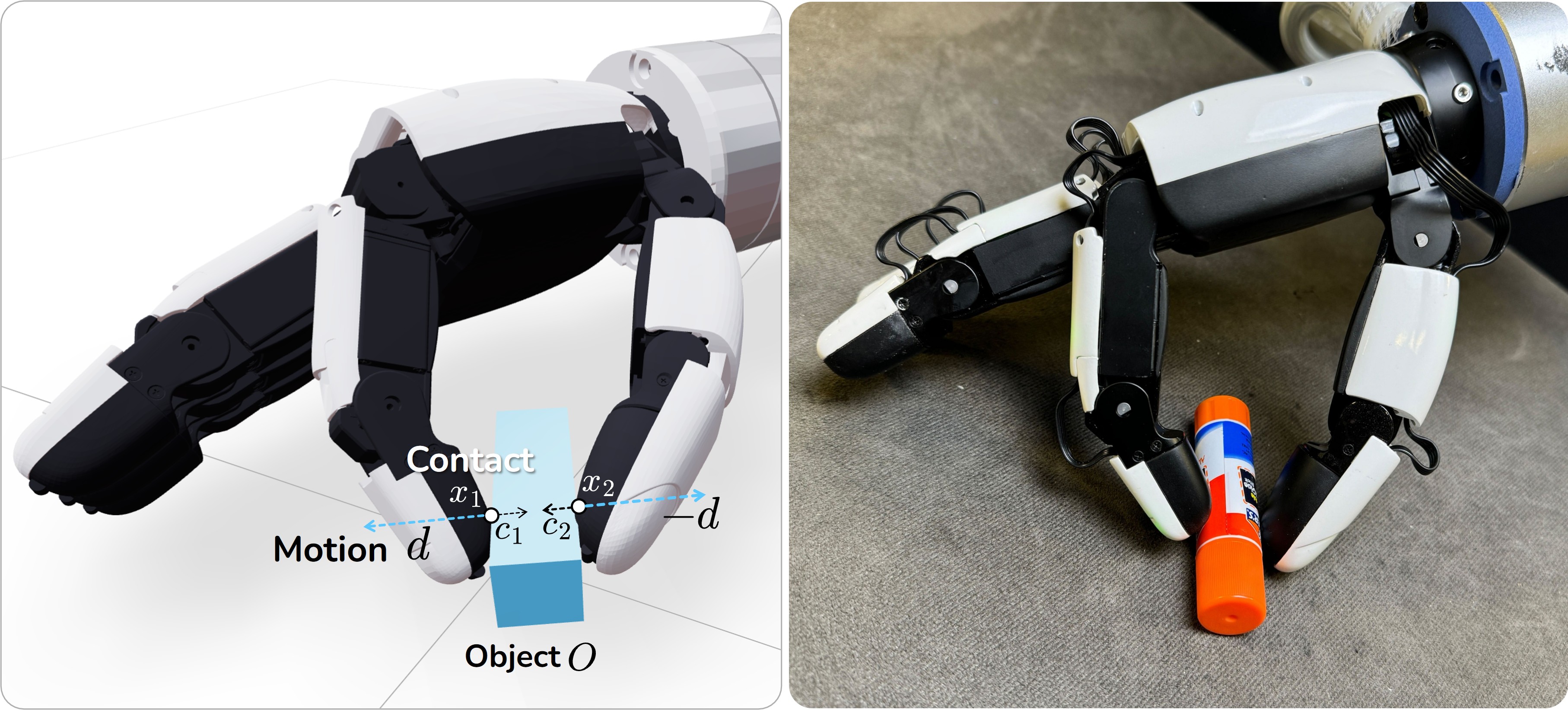}
    \vspace{-1.5em}
    \caption{\textbf{Control Optimization.} For precise grasps control, we optimize for opposite force closure and parallel finger motions, which significantly improve sim-to-real transfer.}
    \label{fig:method_ctrl}
    \vspace{-1.5em}
\end{figure}

After optimizing the grasp pose, a common way for power grasp to obtain pre-grasp and overshoot-grasp poses is based on the object's signed distance function (SDF), which pushes fingers toward the object surface~\cite{ye2025dex1b,wang2023dexgraspnet}. We propose to use simple parallel motions for precision grasps. Let $d$ be the normalized direction vector from the \text{thumb} contact points to the \text{index} contact points. To generate a pre-grasp pose, the \text{thumb} and \text{index} move apart by a small distance $\alpha$. The required joint velocity are calculated using the Jacobian pseudoinverse $J^\dagger$ as $\Delta q_{\text{thumb}} = -\alpha J_{\text{thumb}}^\dagger d$, $\Delta q_{\text{index}} = \alpha J_{\text{index}}^\dagger d$.
Experimental results show that this simple thumb-index grasp with parallel motion is robust for sim-to-real deployment. The finger motion and its real-world counterpart are visualized in Fig.~\ref{fig:method_ctrl}.

All demonstrations are filtered using the ManiSkill simulator~\cite{taomaniskill3}. After data collection, we train two DexSimple~\cite{ye2025dex1b} policies, one on power grasp data and one on precision grasp data. We then add a switcher consisting of PointNet~\cite{qi2017pointnet} and an MLP to predict whether an object should be grasped with a power grasp or a precision grasp, and apply the corresponding policy actions. The policy is deployed to real world in a zero-shot fashion.

\textbf{Real-to-real Tasks.} We tackle both compositional tasks that combine power and precision grasps, and pure precision tasks in a real-to-real setting. Teleoperation is used to collect demonstrations. The standard position-based retargeting~\cite{qin2023anyteleop} struggles with fine-grained actions such as pinching a nut. To address this, we switch between normal retargeting and a pinch controller. The hand pose is optimized by minimizing
$
E_{\mathrm{precise}} = \|G\,c\|_2,
$
where $c$ represents contact normals between the thumb and index finger (rather than between hand and object). We apply the same parallel finger motions used in the sim-to-real setting, where the thumb and index move along direction $d$ with joint updates $\Delta q = J^\dagger d$. Finally, an ACT policy~\cite{zhao2023learning} is trained on these teleoperated demonstrations for deployment.

\subsection{Design Optimization for Precise Manipulation}
\label{subsec:design}

\begin{figure}[t]
    \centering
    \includegraphics[width=0.48\textwidth]{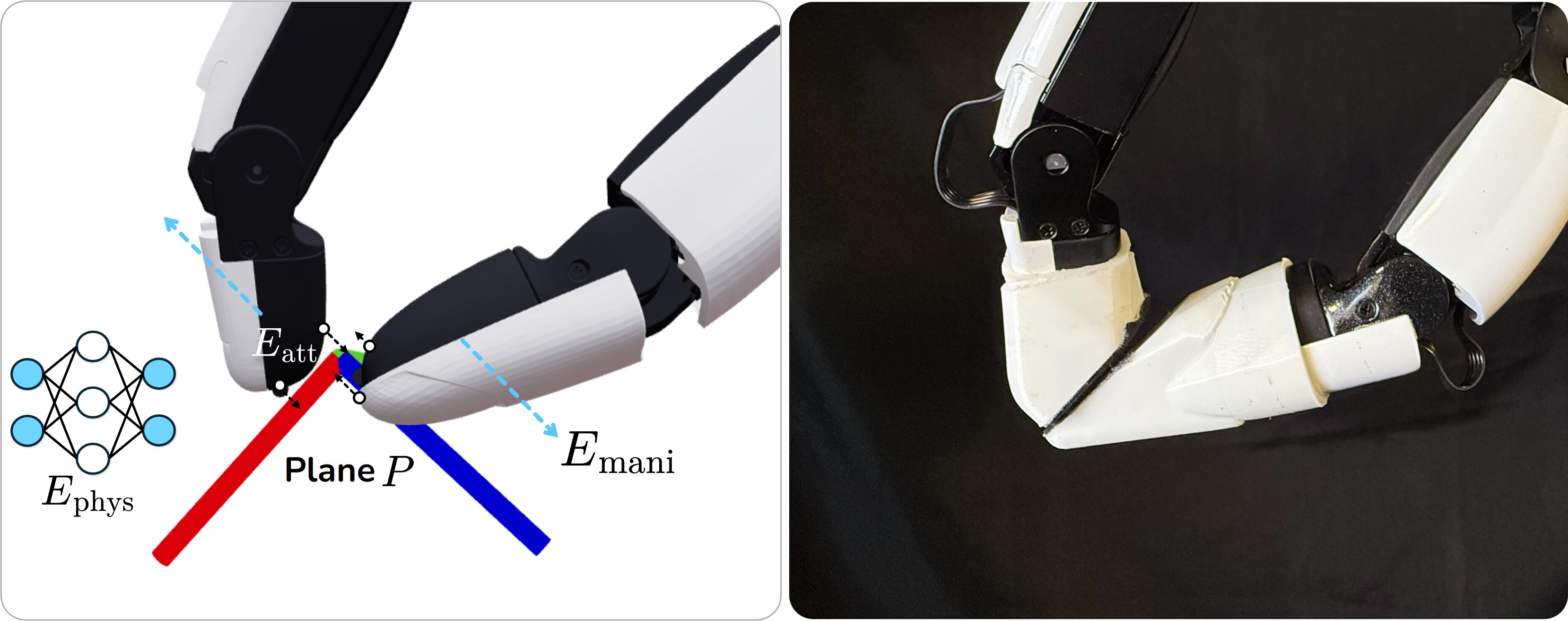}
    \caption{\textbf{Design Optimization.} We optimize fingertip geometry (represented as a contact plane) under multiple objectives. The resulting fingertip cover improves precision manipulation in real-world.}
    \vspace{-1em}
    \label{fig:method_design}
\end{figure}

We aim to optimize hand geometry to enhance precision manipulation without compromising power manipulation.  
To this end, we represent geometry using a simple contact plane $P$, parameterized by a reference point $p$ and a unit normal vector $n$:  $P=\{x\in\mathbb{R}^3 \mid n^\top(x-p)=0\}.$
This simplifies simulation and improves robustness in sim-to-real transfer. Given $P$, we project a slightly inflated convex hull of the fingertip onto it and 3D print the resulting union geometry. The fingertip cover is easy to assemble and generalizes well to different multi-fingered hands.

During plane parameter optimization, we jointly optimize $P$ and the joint position $q$ under multiple objectives using gradient descent. (Note that $q$ here is not used for grasp optimization; it assumes that no object is present.) The objectives and the real-world fingertip covers are shown in Fig.~\ref{fig:method_design}.

\textbf{Geometric Objectives.} The first two objectives encourage thumb-index contact while preventing penetration:  
\begin{equation}
E_{\text{att}}=\sum_{i=1}^{N} d(x_i, P), \quad
E_{\text{rep}}=\sum_{v\in S(F)} [\phi_P(v)<0]\, d(v, P).
\end{equation}  
Here, $x_i$ are candidate contact points on the thumb and index, $S(F)$ is the surface point cloud of finger meshes $F$, $d(\cdot,P)$ is the point-to-plane distance, $\phi_P$ is the signed distance to $P$, and $[\cdot]$ denotes the indicator function. The attraction term pulls sampled contacts toward $P$, while the repulsion term penalizes surface points that cross $P$.

\textbf{Manipulability Objective.}
To further ensure stable motion, we include an objective that encourages the thumb and index to move in parallel along the plane normal direction. We measure this using directional manipulability~\cite{yoshikawa1985manipulability}:  
\begin{equation}
E_{\text{mani}} = -\left( \|J_{\text{thumb}}\,n\|_2 + \|J_{\text{index}}\,n\|_2 \right),
\end{equation}  
where $J_{\text{thumb}}$ and $J_{\text{index}}$ are the Jacobians of the thumb and index, respectively.

\textbf{Neural Physics Objective.}
While the previous terms focus on kinematics and geometry, we further introduce a neural physics objective to leverage large-scale simulation. We first optimize $1$k plane parameters from different initializations using the above objectives, and then evaluate them in simulation on the grasping task introduced in Sec.~\ref{subsec:control}. The simulation outcomes are distilled into a neural surrogate model $f$, implemented as a PointNet encoder followed by an MLP:  
\begin{equation}
f: (P, q, o) \mapsto \hat{s},
\end{equation}
where $(P,q)$ denotes the plane parameters and joint configuration, $o$ is the object observation (point cloud), and $\hat{s}\in [0,1]$ is the predicted task success probability. 
We incorporate this surrogate into the optimization as an energy term:  
\begin{equation}
E_{\text{phys}} = - f(P, q, o),
\end{equation}
which encourages geometries and poses that maximize the predicted success probability. This neural term provides gradient feedback during optimization: we sample a batch of objects and jointly optimize a shared $P$ with diverse $q$. In this way, $P$ is refined toward geometries that are not only kinematically consistent but also physically effective for manipulation.

\section{Experiment}

\subsection{Experimental Settings}

Our method is validated on two platforms: XArm robotic arm + XHand dexterous hand (7+12 DOFs, referred to as xArm below), as well as Unitree G1 Humanoid + Inspire dexterous hand (7+6 DOFs, referred to as G1 below). We cross-validate it on tasks under both sim-to-real and real-to-real settings.

\textbf{Sim-to-real Tasks.} 
We focus on the \textit{grasping} task in the sim-to-real setting, following the Dex1B~\cite{ye2025dex1b} benchmark. 
Our dataset includes 7k Objaverse~\cite{deitke2023objaverse} objects and 1k primitive shapes (spheres, boxes, cylinders) of various sizes. The data are categorized by grasp type.
Objaverse objects with successful poses optimized by $E_\text{precise}$ (Sec.~\ref{subsec:control}) are assigned to precision grasps, while the rest are used for power grasps. All primitive shapes are used for precision grasps. In total, 6k objects are used for power grasps and 3k for precision grasps. We hold out $30\%$ of the objects for testing. We collect 30k trajectories for power and precision grasps, respectively.

The success criterion in Dex1B for grasping is to lift an object from the table to a certain height while maintaining contact between the fingers and the object. During data collection, Dex1B additionally applies lateral external forces to the objects to ensure that the optimized grasp pose is robust, but these external forces are removed during final evaluation. To ensure better sim-to-real transfer, we keep external forces during all evaluations.

We adopt the DexSimple policy~\cite{ye2025dex1b} for the grasping policy. The neural switcher (Sec.~\ref{subsec:control}) consists of a PointNet~\cite{qi2017pointnet} and an MLP, with hidden dimensions (256, 128). The neural physics model (Sec.~\ref{subsec:design}) also consists of a PointNet and an MLP with the same hidden dimensions.

\textbf{Real-to-real Tasks.}
Compared to sim-to-real experiments, we focus on more difficult compositional task and precision task in the real-to-real setting. The task description are as below:

\begin{itemize}
    \item \textbf{Cooking Setup.} A compositional task in which the robot must sequentially pinch-grasp an asparagus spear from the cutting board, place it into the frying pan, then regrasp the pan handle and lift it off the stove.
    \item \textbf{Multi-pen Grasp.} The robot is required to grasp two marker pens and place them into a box within a single attempt. Specifically, it first pinch-grasps one pen and dexterously rolls it into the palm, secured by the remaining three fingers. It then pinch-grasp the second pen, before dropping both pens into the box.
    \item \textbf{Nut onto Peg.} In this task, the robot must precisely pinch-grasp an M4 hex nut (inner diameter $\Phi = 4.1 \,\text{mm}$) from the tabletop and accurately insert it onto an upright M3 bolt (outer diameter $\Phi = 2.9 \,\text{mm}$). The small clearance between the nut and bolt requires fine dexterity and precise alignment, making this task particularly challenging for precise manipulation.
    \item \textbf{Bread Pinch.} In this task, the robot is required to pinch-grasp a single slice of toast from the table. Since excessive downward pressing may deform the bread or even trigger the emergency stop, the robot must execute the grasp with precise control.
    \item \textbf{Battery Insert.} This task involves a sequence of precise manipulations in which the robot must grasp a battery from the table, align it with the charging socket, place it in position, and apply a controlled push to ensure it is fully secured. This task is conducted on G1 setup.
\end{itemize}

\begin{table}[t]
\small
\centering
\setlength{\tabcolsep}{2pt}
\begin{tabular}{lcccccc}
\toprule
& \multicolumn{2}{c}{Optimization} & \multicolumn{2}{c}{Power G. SR (\%)} & \multicolumn{2}{c}{Precise G. SR (\%)} \\
\cmidrule(lr){2-3} \cmidrule(lr){4-5} \cmidrule(lr){6-7}
Method & Design & Control & Seen & Unseen & Seen & Unseen \\
\midrule

\multicolumn{7}{l}{\textbf{\textit{Simulation (xArm)}}} \\
Dex1B~\cite{ye2025dex1b} & & & 59.60 & 55.88 & 56.38 & 53.91 \\
Ours & & $\checkmark$ & 60.64 & 54.12 & 61.54 & 59.04 \\
\rowcolor{myblue}
Ours & $\checkmark$ & $\checkmark$ & \textbf{61.52} & 53.35 & \textbf{64.74} & \textbf{64.17} \\
\midrule

\multicolumn{7}{l}{\textbf{\textit{Simulation (G1)}}} \\
Dex1B~\cite{ye2025dex1b} & & & 60.06 & 56.46 & 44.75 & 44.44 \\
Ours & & $\checkmark$ & 60.54 & 55.42 & 45.62 & 45.26 \\
\rowcolor{myblue}
Ours & $\checkmark$ & $\checkmark$ & 59.93 & \textbf{57.67} & \textbf{49.91} & \textbf{49.32} \\
\midrule

\multicolumn{7}{l}{\textbf{\textit{Zero-shot Sim-to-Real (xArm)}}} \\
Dex1B~\cite{ye2025dex1b} & & & -- & 60.00 & -- & 12.50 \\
Ours & & $\checkmark$ & -- & 70.00 & -- & 20.00 \\
\rowcolor{myblue}
Ours & $\checkmark$ & $\checkmark$ & -- & \textbf{80.00} & -- & \textbf{82.50} \\
\bottomrule
\end{tabular}
\caption{\textbf{Main Results for Sim-to-real Grasping Task.} The top two parts report simulation success rates for G1 and xArm, and the bottom part shows zero-shot sim-to-real results. Our method with joint control and design optimization consistently outperforms Dex1B on precise grasps, especially in real world precision grasps. Power G. and Precise G. denote Power Grasp and Precision Grasp, respectively.}
\vspace{-0.5em}
\label{tab:grasp}
\end{table}

We employ the teleoperation framework from \cite{ding2024bunny} to collect demonstration data, where the teleoperator’s wrist pose is mapped to the XArm end-effector, and the teleoperator’s finger motions are retargeted to the corresponding XHand finger positions. For the baseline retargeting method \cite{qin2023anyteleop}, we adopt the dexpilot retargeting scheme. For the scripted baseline, we record joint configurations from manually executed passive grasping motions. In our optimized retargeting approach, the Euclidean distance between the teleoperator’s thumb and index fingertips is mapped to the opening angle of the corresponding XHand fingers. We collected 15 successful demonstrations for each task.

After data collection, we use \cite{zhao2023learning} for the autonomous policy to verify the policy learning ability in our design. The action trunking transformer (ACT) is implemented by \cite{taomaniskill3}, where the input is third-person camera view with XArm-XHand joint position as proprioception information. The output action is joint targets of Robot Arm and Hand. It is worth noting that we do not employ processed or quantified gripper values for hand supervision, but instead rely solely on raw joint command signals. This aims to validate that our control optimization would not affect the performance of the policy. We tested 15 times for each task on the real robot.    

\subsection{Sim-to-real Results}

\textbf{Main Results.}
The main results of the sim-to-real grasping task are reported in Tab.~\ref{tab:grasp}. We mainly compare against Dex1B~\cite{ye2025dex1b}, a state-of-the-art sim-to-real grasping policy.

\begin{figure}[t]
    \centering
    \includegraphics[width=0.45\textwidth]{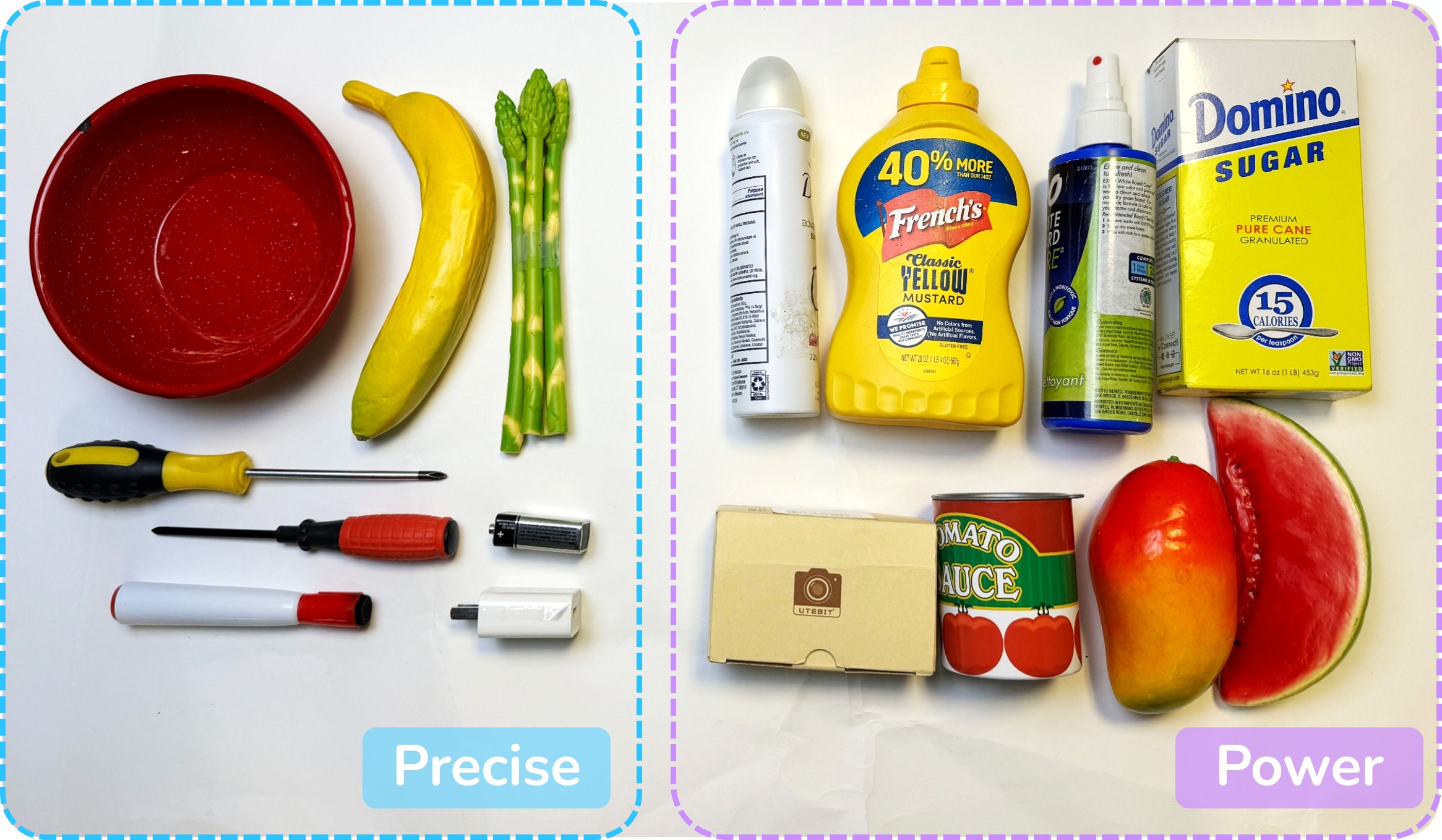}
    \caption{Real-world test objects used for evaluating zero-shot sim-to-real grasping.}
    \vspace{-0.5em}
    \label{fig:test_objects}
\end{figure}

\begin{figure*}[h]
    \centering
    \includegraphics[width=\textwidth]{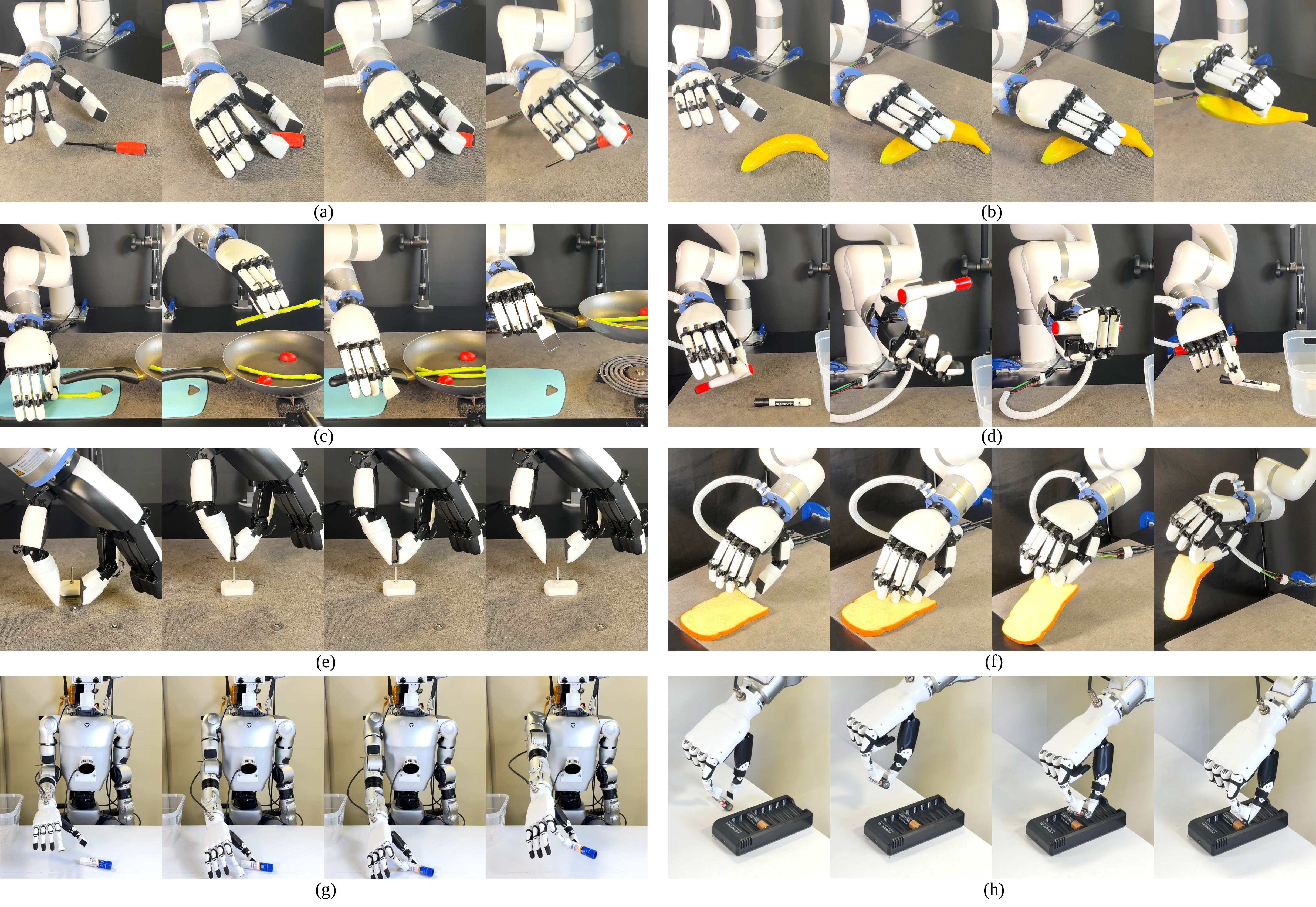}
    \vspace{-1.5em}
    \caption{Task policy rollouts across eight tasks.  (a)-(b) and (g) illustrate sim-to-real transfer of the precision grasp policy on the xArm and G1 setup. (c)-(f) and (h) show real-to-real executions on five distinct tasks: Cooking Setup, Multi-pen Grasp, Nut onto Peg, Bread Pinch, and Battery Insert on xArm and G1.}
    \vspace{-0.5em}
    \label{fig:exp_viz}
\end{figure*}

The simulation results are presented in the top part of Tab.~\ref{tab:grasp}, showing policy success rates for power and precision grasps on both training and testing splits. Both Dex1B and our policy are trained with 30K demonstrations per grasp type. The key differences are that our precision grasp demonstrations are collected using the method described in Sec.~\ref{subsec:control} (Control optimization in the table), and our fingertip geometry is optimized using the method described in Sec.~\ref{subsec:design} (Design optimization in the table). Since all policies are trained on successful demonstrations, the Dex1B policy still achieves strong performance ($53.91\%$) on tiny objects in the precision grasp category. Our policy outperforms Dex1B on precision grasps by about $10\%$ ($64.17\%$ vs $53.91\%$), primarily because parallel finger motions and flat contact planes provide greater robustness when the policy predicts imperfect grasping poses. At the same time, our policy achieves comparable results on power grasps, indicating that the added design optimization does not compromise the dexterous hand’s capability for power grasps.

The zero-shot sim-to-real results are reported in the bottom part of Tab.~\ref{tab:grasp}. The real-world testing objects are shown in Fig.~\ref{fig:test_objects}, and all objects are unseen during training. The policy rollouts are visualized in Fig.~\ref{fig:exp_viz}(a) and (b). We evaluate 5 trials per object, resulting in 40 trials in total for each category. Unlike the simulation results, our method with optimized design outperforms Dex1B by a large margin ($82.50\%$ vs.~$12.50\%$) in real-world deployment. The main reason is that all-finger grasping in Dex1B is often too complex for small objects, making it impractical to deploy under sim-to-real gaps in sensing, calibration, and dynamics. We also observe that our policy with control-only optimization does not achieve strong performance ($20\%$), which indicates that design optimization is crucial for robust sim-to-real transfer, as the flat plane provides significantly more contact area.

\begin{table}[t]
\centering
\small
\setlength{\tabcolsep}{3pt}
\renewcommand{\arraystretch}{1.15}
\begin{tabular}{lcccr}
\toprule
& \multicolumn{2}{c}{Optimization} & \multicolumn{1}{c}{Objective} & \multicolumn{1}{c}{Opt. SR (\%)} \\
\cmidrule(lr){2-3} \cmidrule(lr){4-4}
Method & Design & Control & Physics & \\
\midrule
Dex1B~\cite{ye2025dex1b}   &              &              &              & 2.75 \\
C. only          &              & $\checkmark$ &              & 0.41 \\
C.+D. (w/o $E_\text{phys}$)     & $\checkmark$ & $\checkmark$ &              & 3.77 \\
C.+D.                   & $\checkmark$ & $\checkmark$ & $\checkmark$ & \textbf{5.35} \\
\bottomrule
\end{tabular}
\caption{\textbf{Ablation Study for Precise Grasp Optimization in Simulation.} We report the optimization success rate (Opt. SR; \%) for control-only (C.), joint control and design (C.+D.), and the physics objective $E_\text{phys}$. Joint C.+D. achieves the highest SR ($5.35\%$). Incorporating the $E_\text{phys}$ objective improves performance from $3.77\%$ to $5.35\%$.} 
\label{tab:ablation_precise_opt}
\vspace{-0.5em}
\end{table}

\begin{table*}[t]
\small
\centering
\setlength{\tabcolsep}{6pt}
\begin{tabular}{lccccccc}
\toprule
& \multicolumn{2}{c}{Setting} & \multicolumn{2}{c}{Compositional Task} & \multicolumn{3}{c}{Precision Task} \\
\cmidrule(lr){2-3} \cmidrule(lr){4-5} \cmidrule(lr){6-8}
Method & Design & Control & Cooking Setup & Multi-pen Grasp & Nut onto Peg & Bread Pinch & Battery Insert \\
\midrule
\multicolumn{7}{l}{\textbf{\textit{Autonomous Policy}}} \\
Baseline & Original & Retargeting~\cite{qin2023anyteleop} & 20.0\% & 53.3\% & -- & 60.0\% & 13.3\%\\
\rowcolor{myblue}
Ours & Optimized & Optimized & \textbf{73.3\%} & \textbf{66.7\%} & \textbf{66.7\%} & \textbf{93.3\%} & \textbf{66.7\%} \\
\midrule
\multicolumn{7}{l}{\textbf{\textit{Teleoperation}}} \\
Baseline & Original & Retargeting~\cite{qin2023anyteleop} & 41.7\% & 57.7\% & 6.6\% & 57.1\% & 26.7\% \\
\rowcolor{myblue}
Ours & Optimized & Optimized & \textbf{88.2\%} & \textbf{50.0\%} & \textbf{68.2\%} & \textbf{93.8\%} & \textbf{80.0\%} \\
\bottomrule
\end{tabular}
\caption{Results under autonomous policy and teleoperation on compositional and precision tasks.}
\label{tab:real}
\end{table*}

\textbf{Ablation Study.} We conduct ablation studies in simulation to validate the effectiveness of our proposed techniques. Specifically, we ablate the control optimization, the design optimization, and the $E_\text{phys}$ objective (see Sec.~\ref{sec:method}). Instead of training an imitation learning policy on successful demonstrations and evaluating performance as in Tab.~\ref{tab:grasp}, we directly report the optimization success rates for precise grasp data collection, which serve as a more direct metric of effectiveness.

Our control-only optimization (C. only in the table) achieves a lower success rate ($0.41\%$ vs.~$2.75\%$), since reducing the DOFs makes the optimization problem more challenging. Nevertheless, this control optimization is still valuable because the resulting parallel motions are more robust for learning and sim-to-real transfer. Our joint control and design optimization (C.+D. in the table) achieves the highest success rate ($5.35\%$), highlighting the effectiveness of design optimization. We also evaluate design optimization without the $E_\text{phys}$ objective (C.+D. w/o $E_\text{phys}$ in the table). The performance gap ($3.77\%$ vs.~$5.35\%$) indicates that the $E_\text{phys}$ objective distilled from large-scale simulation helps find more physically plausible fingertip geometries.

\begin{table}[t]
\centering
\small
\setlength{\tabcolsep}{4pt}
\vspace{-0.2em}
\begin{tabular}{llcc}
\toprule
\multicolumn{2}{c}{Setting} & \multicolumn{2}{c}{Success Rate (\%)} \\
\cmidrule(lr){1-2} \cmidrule(lr){3-4}
Design & Control & Policy & Teleoperation \\
\midrule
Original & Retargeting~\cite{qin2023anyteleop} & 60.0\% & 57.1\% \\
Original & Manual Script & 73.3\% & 57.1\% \\
Manual Design & Manual Script & 60.0\% & 50.0\% \\
Optimized (Ours) & Optimized (Ours) & \textbf{93.3\%} & \textbf{93.8\%} \\
\bottomrule
\end{tabular}
\caption{Ablation study on the bread pinch task. We evaluate different combinations of fingertip designs and control methods.}
\vspace{-0.2em}
\label{tab:ablation_bread_pinch}
\end{table}

\subsection{Real-to-real Results}
\textbf{Main Results.} The Table \ref{tab:real} highlights the effectiveness of the proposed co-optimized control and design framework across both autonomous policy and teleoperation settings, for both compositional and precision manipulation task.

In both teleoperation and autonomous settings, the baseline system shows limited success due to the original fingertip design, which makes it difficult to reliably pinch small or thin objects such as asparagus in the Cooking Setup, pens in the Multi-pen Grasp, and thin-cut toast in the Bread Pinch task. With our optimized control–design, success rates increase significantly: for example, from 20.0\% to 73.3\% in Cooking Setup and from 0.0\% to 66.7\% in Nut onto Peg task for the autonomous policies. This demonstrates marked improvements in both compositional and fine-grained precision manipulation skills. Meanwhile, we show that both teleoperation, which maps the operator fingertip distance to the optimized path, and autonomous execution, which learns from joint positions, confirm the effectiveness of our system.

Table \ref{tab:ablation_bread_pinch} reports the ablation study results on the Bread Pinch task. Our optimized fingertip design combined with control consistently outperforms all variants, surpassing both the naive implementation and the manually crafted geometry–control design.

\textbf{Qualitative Results}
To further illustrate the effectiveness of our proposed method, Fig.~\ref{fig:exp_viz} presents task rollouts across eight representative scenarios. Subfigures (a), (b) and (g) demonstrate successful sim-to-real transfer of the precision grasp policy, where the robot reliably manipulates small objects like a screwdriver, banana and glue stick using the co-optimized fingertip design and control strategy.

Subfigures (c)-(f) and (h) highlight real-to-real executions of complex and fine-grained tasks: Cooking Setup, Multi-pen Grasp, Nut onto Peg, Bread Pinch, and Battery Insert. These tasks involve diverse object geometries, contact constraints, dynamic demands and precision demands. In each case, the robot achieves stable and repeatable performance, showcasing the system’s ability to generalize across manipulation contexts. The visual results confirm that the co-optimization approach leads to robust dexterity even in challenging real-world scenarios.

\section{Conclusion}
We introduced a unified framework that enables multi-fingered robotic hands to perform both power and precision grasps through a combination of control-policy learning and fingertip geometry optimization. By simplifying precision control into parallel thumb-index motions and co-designing fingertip covers using a neural-physics surrogate, our method achieves robust, generalizable manipulation without requiring complex hardware modifications.

Extensive evaluations in both sim-to-real and real-world tasks show significant improvements over existing approaches, particularly in fine-grained precision grasping. These results highlight the effectiveness of combining lightweight mechanical design with data-driven control, offering a practical path toward more dexterous and adaptable robotic systems.

\bibliographystyle{IEEEtran}
\bibliography{IEEEabrv,main}

@article{ye2023learning,
  title={Learning continuous grasping function with a dexterous hand from human demonstrations},
  author={Ye, Jianglong and Wang, Jiashun and Huang, Binghao and Qin, Yuzhe and Wang, Xiaolong},
  journal={RA-L},
  year={2023}
}

@inproceedings{zhang2024graspxl,
  title={Graspxl: Generating grasping motions for diverse objects at scale},
  author={Zhang, Hui and Christen, Sammy and Fan, Zicong and Hilliges, Otmar and Song, Jie},
  booktitle={ECCV},
  year={2024},
}

@inproceedings{ye2025dex1b,
  title={Dex1B: Learning with 1B Demonstrations for Dexterous Manipulation},
  author={Ye, Jianglong and Wang, Keyi and Yuan, Chengjing and Yang, Ruihan and Li, Yiquan and Zhu, Jiyue and Qin, Yuzhe and Zou, Xueyan and Wang, Xiaolong},
  booktitle={Robotics: Science and Systems (RSS)},
  year={2025}
}

@inproceedings{qin2023anyteleop,
  title     = {AnyTeleop: A General Vision-Based Dexterous Robot Arm-Hand Teleoperation System},
  author    = {Qin, Yuzhe and Yang, Wei and Huang, Binghao and Van Wyk, Karl and Su, Hao and Wang, Xiaolong and Chao, Yu-Wei and Fox, Dieter},
  booktitle = {Robotics: Science and Systems (RSS)},
  year      = {2023}
}

@article{napier1956prehensile,
  title={The prehensile movements of the human hand.},
  author={Napier, John R},
  journal={The Journal of Bone and Joint Surgery. British volume},
  year={1956},
}

@article{feix2015grasp,
  title={The grasp taxonomy of human grasp types},
  author={Feix, Thomas and Romero, Javier and Schmiedmayer, Heinz-Bodo and Dollar, Aaron M and Kragic, Danica},
  journal={Transactions on Human-Machine Systems},
  year={2015},
}

@article{skinner2015human,
  title={Human-like hand use in Australopithecus africanus},
  author={Skinner, Matthew M and Stephens, Nicholas B and Tsegai, Zewdi J and Foote, Alexandra C and Nguyen, N Huynh and Gross, Thomas and Pahr, Dieter H and Hublin, Jean-Jacques and Kivell, Tracy L},
  journal={Science},
  year={2015},
}

@article{kivell2015evidence,
  title={Evidence in hand: recent discoveries and the early evolution of human manual manipulation},
  author={Kivell, Tracy L},
  journal={Philosophical Transactions of the Royal Society B: Biological Sciences},
  year={2015},
}

@article{karakostis2021biomechanics,
  title={Biomechanics of the human thumb and the evolution of dexterity},
  author={Karakostis, Fotios Alexandros and Haeufle, Daniel and Anastopoulou, Ioanna and Moraitis, Konstantinos and Hotz, Gerhard and Tourloukis, Vangelis and Harvati, Katerina},
  journal={Current Biology},
  year={2021},
}

@inproceedings{chen2025bodex,
  title={Bodex: Scalable and efficient robotic dexterous grasp synthesis using bilevel optimization},
  author={Chen, Jiayi and Ke, Yubin and Wang, He},
  booktitle={International Conference on Robotics and Automation (ICRA)},
  year={2025},
}

@article{fang2025anydexgrasp,
  title={AnyDexGrasp: General Dexterous Grasping for Different Hands with Human-level Learning Efficiency},
  author={Fang, Hao-Shu and Yan, Hengxu and Tang, Zhenyu and Fang, Hongjie and Wang, Chenxi and Lu, Cewu},
  journal={arXiv preprint arXiv:2502.16420},
  year={2025}
}

@article{intelligence2025pi05,
  title={$\pi$0.5: a Vision-Language-Action Model with Open-World Generalization},
  author={Intelligence, Physical and Black, Kevin and Brown, Noah and Darpinian, James and Dhabalia, Karan and Driess, Danny and Esmail, Adnan and Equi, Michael and Finn, Chelsea and Fusai, Niccolo and others},
  journal={ArXiv},
  year={2025},
}

@article{team2025gemini,
  title={Gemini robotics: Bringing ai into the physical world},
  author={Team, Gemini Robotics and Abeyruwan, Saminda and Ainslie, Joshua and Alayrac, Jean-Baptiste and Arenas, Montserrat Gonzalez and Armstrong, Travis and Balakrishna, Ashwin and Baruch, Robert and Bauza, Maria and Blokzijl, Michiel and others},
  journal={arXiv preprint arXiv:2503.20020},
  year={2025}
}

@article{yoshikawa1985manipulability,
  title={Manipulability of robotic mechanisms},
  author={Yoshikawa, Tsuneo},
  journal={The international journal of Robotics Research},
  year={1985},
}

@article{ding2024bunny,
  title={Bunny-visionpro: Real-time bimanual dexterous teleoperation for imitation learning},
  author={Ding, Runyu and Qin, Yuzhe and Zhu, Jiyue and Jia, Chengzhe and Yang, Shiqi and Yang, Ruihan and Qi, Xiaojuan and Wang, Xiaolong},
  journal={arXiv preprint arXiv:2407.03162},
  year={2024}
}

@article{chen2020design,
  title={Design optimization of soft robots: A review of the state of the art},
  author={Chen, Feifei and Wang, Michael Yu},
  journal={IEEE Robotics \& Automation Magazine},
  volume={27},
  number={4},
  pages={27--43},
  year={2020},
  publisher={IEEE}
}

@article{calandra2016bayesian,
  title={Bayesian optimization for learning gaits under uncertainty: An experimental comparison on a dynamic bipedal walker},
  author={Calandra, Roberto and Seyfarth, Andr{\'e} and Peters, Jan and Deisenroth, Marc Peter},
  journal={Annals of Mathematics and Artificial Intelligence},
  volume={76},
  pages={5--23},
  year={2016},
  publisher={Springer}
}

@inproceedings{feshbach2024algorithmic,
  title={Algorithmic design of kinematic trees based on CSC Dubins planning for link shapes},
  author={Feshbach, Daniel and Chen, Wei-Hsi and Xu, Ling and Schaumburg, Emil and Huang, Isabella and Sung, Cynthia},
  booktitle={International Workshop on the Algorithmic Foundations of Robotics (WAFR)},
  year={2024}
}

@inproceedings{wang2023dexgraspnet,
  title={Dexgraspnet: A large-scale robotic dexterous grasp dataset for general objects based on simulation},
  author={Wang, Ruicheng and Zhang, Jialiang and Chen, Jiayi and Xu, Yinzhen and Li, Puhao and Liu, Tengyu and Wang, He},
  booktitle={ICRA},
  year={2023},
}

@article{zhao2023learning,
  title={Learning fine-grained bimanual manipulation with low-cost hardware},
  author={Zhao, Tony Z and Kumar, Vikash and Levine, Sergey and Finn, Chelsea},
  journal={RSS},
  year={2023}
}

@article{taomaniskill3,
  title={ManiSkill3: GPU Parallelized Robotics Simulation and Rendering for Generalizable Embodied AI},
  author={Stone Tao and Fanbo Xiang and Arth Shukla and Yuzhe Qin and Xander Hinrichsen and Xiaodi Yuan and Chen Bao and Xinsong Lin and Yulin Liu and Tse-kai Chan and Yuan Gao and Xuanlin Li and Tongzhou Mu and Nan Xiao and Arnav Gurha and Zhiao Huang and Roberto Calandra and Rui Chen and Shan Luo and Hao Su},
  journal = {arXiv},
  year={2024},
}

@inproceedings{qi2017pointnet,
  title={Pointnet: Deep learning on point sets for 3d classification and segmentation},
  author={Qi, Charles R and Su, Hao and Mo, Kaichun and Guibas, Leonidas J},
  booktitle={CVPR},
  year={2017}
}

@article{liu2021synthesizing,
  title={Synthesizing diverse and physically stable grasps with arbitrary hand structures using differentiable force closure estimator},
  author={Liu, Tengyu and Liu, Zeyu and Jiao, Ziyuan and Zhu, Yixin and Zhu, Song-Chun},
  journal={RA-L},
  year={2021},
}

@inproceedings{zhao2024aloha,
  title={Aloha unleashed: A simple recipe for robot dexterity},
  author={Zhao, Tony Z and Tompson, Jonathan and Driess, Danny and Florence, Pete and Ghasemipour, Kamyar and Finn, Chelsea and Wahid, Ayzaan},
  booktitle={CoRL},
  year={2024}
}

@article{black2024pi_0,
  title={{$\pi 0$}: A Vision-Language-Action Flow Model for General Robot Control},
  author={Black, Kevin and Brown, Noah and Driess, Danny and Esmail, Adnan and Equi, Michael and Finn, Chelsea and Fusai, Niccolo and Groom, Lachy and Hausman, Karol and Ichter, Brian and others},
  journal={arXiv},
  year={2024}
}

@inproceedings{qin2022dexmv,
  title={Dexmv: Imitation learning for dexterous manipulation from human videos},
  author={Qin, Yuzhe and Wu, Yueh-Hua and Liu, Shaowei and Jiang, Hanwen and Yang, Ruihan and Fu, Yang and Wang, Xiaolong},
  booktitle={ECCV},
  year={2022},
}

@article{singh2024dextrah,
  title={DextrAH-RGB: Visuomotor Policies to Grasp Anything with Dexterous Hands},
  author={Singh, Ritvik and Allshire, Arthur and Handa, Ankur and Ratliff, Nathan and Van Wyk, Karl},
  journal={arXiv preprint arXiv:2412.01791},
  year={2024}
}

@article{he2025dexvlg,
  title={DexVLG: Dexterous Vision-Language-Grasp Model at Scale},
  author={He, Jiawei and Li, Danshi and Yu, Xinqiang and Qi, Zekun and Zhang, Wenyao and Chen, Jiayi and Zhang, Zhaoxiang and Zhang, Zhizheng and Yi, Li and Wang, He},
  journal={arXiv preprint arXiv:2507.02747},
  year={2025}
}

@article{zhong2025dexgraspvla,
  title={Dexgraspvla: A vision-language-action framework towards general dexterous grasping},
  author={Zhong, Yifan and Huang, Xuchuan and Li, Ruochong and Zhang, Ceyao and Chen, Zhang and Guan, Tianrui and Zeng, Fanlian and Lui, Ka Num and Ye, Yuyao and Liang, Yitao and others},
  journal={arXiv preprint arXiv:2502.20900},
  year={2025}
}

@article{zhao2023gelsight,
  title={Gelsight svelte hand: A three-finger, two-dof, tactile-rich, low-cost robot hand for dexterous manipulation},
  author={Zhao, Jialiang and Adelson, Edward H},
  journal={arXiv preprint arXiv:2309.10886},
  year={2023}
}

@inproceedings{finn2017one,
  title={One-shot visual imitation learning via meta-learning},
  author={Finn, Chelsea and Yu, Tianhe and Zhang, Tianhao and Abbeel, Pieter and Levine, Sergey},
  booktitle={Conference on robot learning},
  pages={357--368},
  year={2017},
  organization={PMLR}
}

@article{mandlekar2020learning,
  title={Learning to generalize across long-horizon tasks from human demonstrations},
  author={Mandlekar, Ajay and Xu, Danfei and Mart{\'\i}n-Mart{\'\i}n, Roberto and Savarese, Silvio and Fei-Fei, Li},
  journal={arXiv preprint arXiv:2003.06085},
  year={2020}
}

@article{zhu2022bottom,
  title={Bottom-up skill discovery from unsegmented demonstrations for long-horizon robot manipulation},
  author={Zhu, Yifeng and Stone, Peter and Zhu, Yuke},
  journal={IEEE Robotics and Automation Letters},
  volume={7},
  number={2},
  pages={4126--4133},
  year={2022},
  publisher={IEEE}
}

@inproceedings{wu2024gello,
  title={Gello: A general, low-cost, and intuitive teleoperation framework for robot manipulators},
  author={Wu, Philipp and Shentu, Yide and Yi, Zhongke and Lin, Xingyu and Abbeel, Pieter},
  booktitle={IROS},
  year={2024},
}

@inproceedings{young2021visual,
  title={Visual imitation made easy},
  author={Young, Sarah and Gandhi, Dhiraj and Tulsiani, Shubham and Gupta, Abhinav and Abbeel, Pieter and Pinto, Lerrel},
  booktitle={Conference on Robot learning},
  pages={1992--2005},
  year={2021},
  organization={PMLR}
}

@inproceedings{mandlekar2018roboturk,
  title={Roboturk: A crowdsourcing platform for robotic skill learning through imitation},
  author={Mandlekar, Ajay and Zhu, Yuke and Garg, Animesh and Booher, Jonathan and Spero, Max and Tung, Albert and Gao, Julian and Emmons, John and Gupta, Anchit and Orbay, Emre and others},
  booktitle={Conference on Robot Learning},
  pages={879--893},
  year={2018},
  organization={PMLR}
}

@article{wang2023robogen,
  title={Robogen: Towards unleashing infinite data for automated robot learning via generative simulation},
  author={Wang, Yufei and Xian, Zhou and Chen, Feng and Wang, Tsun-Hsuan and Wang, Yian and Fragkiadaki, Katerina and Erickson, Zackory and Held, David and Gan, Chuang},
  journal={arXiv preprint arXiv:2311.01455},
  year={2023}
}

@article{mandlekar2021matters,
  title={What matters in learning from offline human demonstrations for robot manipulation},
  author={Mandlekar, Ajay and Xu, Danfei and Wong, Josiah and Nasiriany, Soroush and Wang, Chen and Kulkarni, Rohun and Fei-Fei, Li and Savarese, Silvio and Zhu, Yuke and Mart{\'\i}n-Mart{\'\i}n, Roberto},
  journal={arXiv preprint arXiv:2108.03298},
  year={2021}
}

@article{mandlekar2023mimicgen,
  title={Mimicgen: A data generation system for scalable robot learning using human demonstrations},
  author={Mandlekar, Ajay and Nasiriany, Soroush and Wen, Bowen and Akinola, Iretiayo and Narang, Yashraj and Fan, Linxi and Zhu, Yuke and Fox, Dieter},
  journal={arXiv preprint arXiv:2310.17596},
  year={2023}
}

@article{james2020rlbench,
  title={Rlbench: The robot learning benchmark \& learning environment},
  author={James, Stephen and Ma, Zicong and Arrojo, David Rovick and Davison, Andrew J},
  journal={IEEE Robotics and Automation Letters},
  volume={5},
  number={2},
  pages={3019--3026},
  year={2020},
  publisher={IEEE}
}

@article{nasiriany2022learning,
  title={Learning and retrieval from prior data for skill-based imitation learning},
  author={Nasiriany, Soroush and Gao, Tian and Mandlekar, Ajay and Zhu, Yuke},
  journal={arXiv preprint arXiv:2210.11435},
  year={2022}
}

@inproceedings{kalashnikov2018scalable,
  title={Scalable deep reinforcement learning for vision-based robotic manipulation},
  author={Kalashnikov, Dmitry and Irpan, Alex and Pastor, Peter and Ibarz, Julian and Herzog, Alexander and Jang, Eric and Quillen, Deirdre and Holly, Ethan and Kalakrishnan, Mrinal and Vanhoucke, Vincent and others},
  booktitle={Conference on robot learning},
  pages={651--673},
  year={2018},
  organization={PMLR}
}

@inproceedings{o2024open,
  title={Open x-embodiment: Robotic learning datasets and rt-x models: Open x-embodiment collaboration 0},
  author={O’Neill, Abby and Rehman, Abdul and Maddukuri, Abhiram and Gupta, Abhishek and Padalkar, Abhishek and Lee, Abraham and Pooley, Acorn and Gupta, Agrim and Mandlekar, Ajay and Jain, Ajinkya and others},
  booktitle={2024 IEEE International Conference on Robotics and Automation (ICRA)},
  pages={6892--6903},
  year={2024},
  organization={IEEE}
}

@inproceedings{bahl2023affordances,
  title={Affordances from human videos as a versatile representation for robotics},
  author={Bahl, Shikhar and Mendonca, Russell and Chen, Lili and Jain, Unnat and Pathak, Deepak},
  booktitle={Proceedings of the IEEE/CVF Conference on Computer Vision and Pattern Recognition},
  pages={13778--13790},
  year={2023}
}

@article{dass2022pato,
  title={Pato: Policy assisted teleoperation for scalable robot data collection},
  author={Dass, Shivin and Pertsch, Karl and Zhang, Hejia and Lee, Youngwoon and Lim, Joseph J and Nikolaidis, Stefanos},
  journal={arXiv preprint arXiv:2212.04708},
  year={2022}
}

@inproceedings{pinto2016supersizing,
  title={Supersizing self-supervision: Learning to grasp from 50k tries and 700 robot hours},
  author={Pinto, Lerrel and Gupta, Abhinav},
  booktitle={2016 IEEE international conference on robotics and automation (ICRA)},
  pages={3406--3413},
  year={2016},
  organization={IEEE}
}

@article{hong2023angle,
  title={Angle-programmed tendril-like trajectories enable a multifunctional gripper with ultradelicacy, ultrastrength, and ultraprecision},
  author={Hong, Yaoye and Zhao, Yao and Berman, Joseph and Chi, Yinding and Li, Yanbin and Huang, He and Yin, Jie},
  journal={Nature Communications},
  volume={14},
  number={1},
  pages={4625},
  year={2023},
  publisher={Nature Publishing Group UK London}
}

@article{brohan2022rt,
  title={Rt-1: Robotics transformer for real-world control at scale},
  author={Brohan, Anthony and Brown, Noah and Carbajal, Justice and Chebotar, Yevgen and Dabis, Joseph and Finn, Chelsea and Gopalakrishnan, Keerthana and Hausman, Karol and Herzog, Alex and Hsu, Jasmine and others},
  journal={arXiv preprint arXiv:2212.06817},
  year={2022}
}

@inproceedings{do2024densetact,
  title={Densetact-mini: an optical tactile sensor for grasping multi-scale objects from flat surfaces},
  author={Do, Won Kyung and Dhawan, Ankush Kundan and Kitzmann, Mathilda and Kennedy, Monroe},
  booktitle={2024 IEEE international conference on robotics and automation (ICRA)},
  pages={6928--6934},
  year={2024},
  organization={IEEE}
}

@inproceedings{yu2023precise,
  title={Precise robotic needle-threading with tactile perception and reinforcement learning},
  author={Yu, Zhenjun and Xu, Wenqiang and Yao, Siqiong and Ren, Jieji and Tang, Tutian and Li, Yutong and Gu, Guoying and Lu, Cewu},
  booktitle={Conference on Robot Learning},
  pages={3266--3276},
  year={2023},
  organization={PMLR}
}

@inproceedings{bronars2024texterity,
  title={Texterity: Tactile extrinsic dexterity},
  author={Bronars, Antonia and Kim, Sangwoon and Patre, Parag and Rodriguez, Alberto},
  booktitle={2024 IEEE International Conference on Robotics and Automation (ICRA)},
  pages={7976--7983},
  year={2024},
  organization={IEEE}
}

@inproceedings{mannam2024design,
  title={Design and control co-optimization for automated design iteration of dexterous anthropomorphic soft robotic hands},
  author={Mannam, Pragna and Liu, Xingyu and Zhao, Ding and Oh, Jean and Pollard, Nancy},
  booktitle={2024 IEEE 7th International Conference on Soft Robotics (RoboSoft)},
  pages={332--339},
  year={2024},
  organization={IEEE}
}

@inproceedings{ikemuraefficient,
  title={Efficient End-effector Co-Design by Demonstration for Deformable Fragile Object Manipulation},
  author={Ikemura, Kei and Dong, Yifei and Blanco-Mulero, David and Longhini, Alberta and Chen, Li and Pokorny, Florian T},
  booktitle={1st Workshop on Robot Hardware-Aware Intelligence}
}

@article{liu2024paperbot,
  title={Paperbot: Learning to design real-world tools using paper},
  author={Liu, Ruoshi and Liang, Junbang and Sudhakar, Sruthi and Ha, Huy and Chi, Cheng and Song, Shuran and Vondrick, Carl},
  journal={arXiv preprint arXiv:2403.09566},
  year={2024}
}

@article{yi2025co,
  title={Co-Design of Soft Gripper with Neural Physics},
  author={Yi, Sha and Bai, Xueqian and Singh, Adabhav and Ye, Jianglong and Tolley, Michael T and Wang, Xiaolong},
  journal={arXiv preprint arXiv:2505.20404},
  year={2025}
}

@article{cheng2024open,
  title={Open-television: Teleoperation with immersive active visual feedback},
  author={Cheng, Xuxin and Li, Jialong and Yang, Shiqi and Yang, Ge and Wang, Xiaolong},
  journal={arXiv preprint arXiv:2407.01512},
  year={2024}
}

@inproceedings{deitke2023objaverse,
  title={Objaverse: A universe of annotated 3d objects},
  author={Deitke, Matt and Schwenk, Dustin and Salvador, Jordi and Weihs, Luca and Michel, Oscar and VanderBilt, Eli and Schmidt, Ludwig and Ehsani, Kiana and Kembhavi, Aniruddha and Farhadi, Ali},
  booktitle={Proceedings of the IEEE/CVF conference on computer vision and pattern recognition},
  pages={13142--13153},
  year={2023}
}

@article{wei2024ensuringforcesafetyvisionguided,
      title={Ensuring Force Safety in Vision-Guided Robotic Manipulation via Implicit Tactile Calibration}, 
      author={Lai Wei and Jiahua Ma and Yibo Hu and Ruimao Zhang},
      journal={arXiv preprint arXiv:2412.10349},
      year={2024},
}

@article{qiu-song-peng-2024-wildlma,
    title={WildLMa: Long Horizon Loco-Manipulation in the Wild},
    author={Ri-Zhao Qiu and Yuchen Song and Xuanbin Peng and Sai Aneesh Suryadevara and Ge Yang and Minghuan Liu and Mazeyu Ji and Chengzhe Jia and Ruihan Yang and Xueyan Zou and Xiaolong Wang},    journal={arXiv preprint arXiv:2411.15131},
    year={2024}
}

\end{document}